\documentclass{article}
\usepackage{ijcai20}

\usepackage{times}
\usepackage{soul}
\usepackage{url}
\usepackage[hidelinks]{hyperref}
\usepackage[utf8]{inputenc}
\usepackage[small]{caption}
\usepackage{graphicx}
\usepackage{amsmath}
\usepackage{amsthm}
\usepackage{booktabs}
\usepackage{algorithm}
\usepackage{algorithmic}
\usepackage{amsfonts}
\usepackage{mathabx}
\usepackage{xcolor}
\usepackage[capitalize]{cleveref}
\usepackage{enumitem}
\usepackage{multirow}

\usepackage[mathscr]{euscript}

\let\vec\mathbf

\title{GraphFlow: Exploiting Conversation Flow with Graph Neural Networks for Conversational Machine Comprehension}

\author{
Yu Chen$^1$
\and
Lingfei Wu$^2$ \thanks{Corresponding author.} \And
Mohammed J. Zaki$^{1}$
\affiliations
$^1$Rensselaer Polytechnic Institute\\
$^2$IBM Research
\emails
cheny39@rpi.edu,
lwu@email.wm.edu,
zaki@cs.rpi.edu
}

\begin{document}

\maketitle

\begin{abstract}

Conversational machine comprehension (MC) has proven significantly more challenging compared to traditional MC since it requires better utilization of conversation history. However, most existing approaches do not effectively capture conversation history and thus have trouble handling questions involving coreference or ellipsis. Moreover, when reasoning over passage text, most of them simply treat it as a word sequence without exploring rich semantic relationships among words. In this paper, we first propose a simple yet effective graph structure learning technique to dynamically construct a question and conversation history aware context graph at each conversation turn. Then we propose a novel Recurrent Graph Neural Network, and based on that, we introduce a flow mechanism to model the temporal dependencies in a sequence of context graphs. The proposed \textsc{GraphFlow} model can effectively capture conversational flow in a dialog, and shows competitive performance compared to existing state-of-the-art methods on CoQA, QuAC and DoQA benchmarks. In addition, visualization experiments show that our proposed model can offer good interpretability for the reasoning process.


\end{abstract}

\section{Introduction}

Recent years have observed a surge of interest in conversational machine comprehension (MC).
Unlike the setting of traditional MC that requires answering a single question given a passage, the conversational MC task is to answer a question in a conversation given a passage and all previous questions and answers.
Despite the success existing works have achieved on MC (e.g., SQuAD \cite{rajpurkar2016squad}), conversational MC has proven significantly more challenging.
We highlight two major challenges here.
First, the focus usually shifts as a conversation progresses~\cite{reddy2018coqa,choi2018quac}.
Second, many questions refer back to conversation history via coreference or ellipsis.
Therefore, without fully utilizing conversation history (i.e., previous questions and/or answers), one can not understand a question correctly.
In this work, we model the concept of \textsl{conversation flow} as a sequence of latent states associated with these shifts of focus in a conversation.

To cope with the above challenges,
many methods have been proposed to utilize conversation history.
Most of them simply prepend the conversation history to a current question \cite{reddy2018coqa,zhu2018sdnet} or add previous answer locations to a passage \cite{choi2018quac,yatskar2018qualitative}, and then treat the task as single-turn MC
without explicitly modeling conversation flow.
Huang et al.~\shortcite{huang2018flowqa} assumed that hidden representations generated during previous reasoning processes potentially capture important information for answering a current question.
In order to model conversation flow,
they proposed an \textsl{Integration-Flow} (IF) mechanism to first perform sequential reasoning over passage words in parallel for each turn, and then refine the reasoning results sequentially across different turns, in parallel of passage words.

However, the IF mechanism has several limitations when reasoning over a sequence of passages for answer seeking.
First of all, the strategy of interweaving two processing directions (i.e., in passage words and in question turns) is not quite effective.
Because in the IF mechanism, the results of previous reasoning processes are not incorporated into the current reasoning process immediately.
Instead, all reasoning processes over passage words are conducted in parallel.
As a result, the reasoning performance at each turn is not improved by the outcome of previous reasoning processes.
To alleviate this issue, they have to refine the reasoning results sequentially across different turns and use stacked IF layers to interweave two processing directions multiple times.
Second, following most previous methods,
when reasoning over passage text, they simply treat it as a word sequence without exploring the rich semantic relationships among words.
Recent works on multi-hop MC~\cite{de2018question,song2018exploring} have shown the advantages of applying a Graph Neural Network (GNN) to process a passage graph over simply processing a word sequence using a Recurrent Neural Network (RNN).

To better capture conversation flow and address the above issues, in this work,
we propose \textsc{GraphFlow}, a GNN based model for conversational MC.
We first propose a simple yet effective graph structure learning technique to dynamically construct a question and conversation history aware context graph at each turn that consists of each word as a node.
Then we propose a novel Recurrent Graph Neural Network (RGNN), and based on that, we introduce a flow mechanism to model the temporal dependencies in a sequence of context graphs.
Answers are finally predicted based on the matching score of the question embedding and the context graph embedding at each turn.

We highlight our contributions as follows:
\setlist{nolistsep}
\begin{itemize}[noitemsep]

    \item We propose a novel GNN based model, namely \textsc{GraphFlow}, for conversational MC which captures conversational flow in a dialog.
    \item We dynamically construct a question and conversation history aware context graph at each turn, and propose a novel Recurrent Graph Neural Network based flow mechanism to process a sequence of context graphs.
    \item On three public benchmarks (i.e., CoQA, QuAC and DoQA), our model shows competitive performance compared to existing state-of-the-art methods. In addition, visualization experiments show that our model can offer good interpretability for the reasoning process.
\end{itemize}


\section{Related Work}
\subsection{Conversational MC}\label{sec:related_work_crc}
One big challenge of Conversational MC is how to effectively utilize conversation history.
\cite{reddy2018coqa,zhu2018sdnet} concatenated previous questions and answers to the current question.
Choi et al.~\shortcite{choi2018quac}
concatenated a feature vector encoding the turn number to the question word embedding and a feature vector encoding previous N answer locations to the context embeddings.
However, these methods ignore previous reasoning processes performed by the model when reasoning at the current turn.
Huang et al.~\shortcite{huang2018flowqa} proposed the idea of \textsl{Integration-Flow} (IF) to allow rich information in the reasoning process to flow through a conversation.
To better model conversation flow,
in this work, we propose a novel GNN based flow mechanism to sequentially process a sequence of context graphs.

Another challenge of this task is how to handle abstractive answers.
Reddy et al.~\shortcite{reddy2018coqa}
propose a hybrid method DrQA+PGNet, which augments a traditional extractive reading comprehension model with a text generator.
Yatskar~\shortcite{yatskar2018qualitative}
propose to first make a Yes/No decision, and output an answer span only if Yes/No was not selected.
Recent work \cite{huang2018flowqa,zhu2018sdnet,yeh2019flowdelta,qu2019attentive,ju2019technical} as well as our work in this paper follows a similar idea to handle abstractive answers.

When processing passage text in MC, most existing methods treat it as a word sequence. Recently, promising results have been achieved by applying a GNN to process a passage graph~\cite{de2018question,song2018exploring}.

\subsection{Graph Neural Networks}

Over the past few years, graph neural networks (GNNs) \cite{kipf2016semi,gilmer2017neural,hamilton2017inductive,xu2018graph2seq} have drawn increasing attention.
Recently, GNNs have been applied to various question answering tasks including
knowledge base question answering (KBQA) \cite{sun2018open}, question generation \cite{chen2020toward}, and MC \cite{de2018question,song2018exploring}, and have shown advantages over traditional approaches.
For tasks where the graph structure is unknown,
linguistic features (e.g., dependency parsing, coreferences)~\cite{xu2018exploiting,de2018question,song2018exploring} or attention-based mechanisms~\cite{liu2018contextualized,chen2019reinforcement,chen2019deep}
are usually used to construct a static or dynamic graph containing words or entity mentions as nodes.

\section{The GraphFlow Approach}
The task of conversational MC is to answer a natural language question given the context and conversation history.
Let us denote $C$ as the context which consists of a word sequence $\{c_1, c_2, ..., c_m\}$
and $Q^{(i)}$ as the question at the $i$-th turn which consists of a word sequence $\{q_{1}^{(i)}, q_{2}^{(i)}, ..., q_{n}^{(i)}\}$.
And there are totally $T$ turns in a conversation.

\begin{figure}
  \centering
    \includegraphics[
    keepaspectratio=true,scale=0.178
    ]{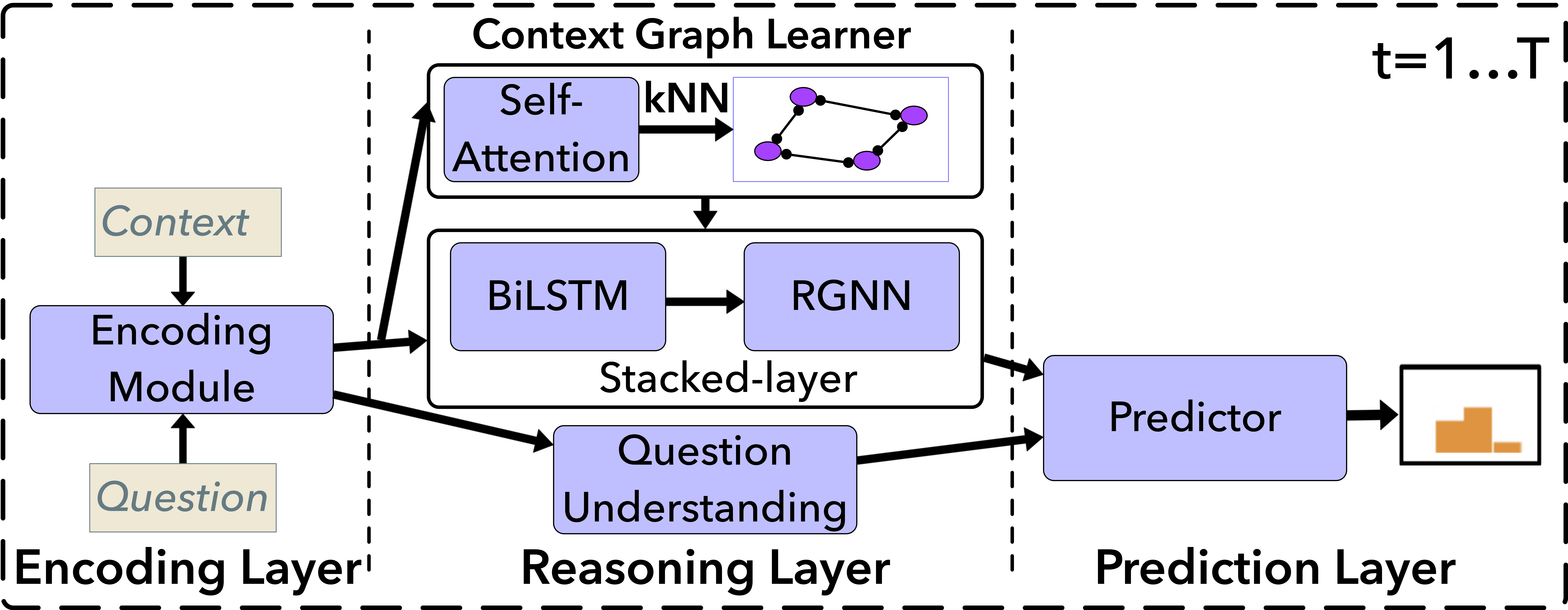}
  \caption{Overall architecture of the proposed model. }
  \label{fig:overall_arch}
\end{figure}

As shown in \cref{fig:overall_arch},
our proposed \textsc{GraphFlow} model
consists of {\em Encoding Layer}, {\em Reasoning Layer} and {\em Prediction Layer}.
The {\em Encoding Layer} encodes conversation history and context that aligns question information.
The {\em Reasoning Layer} dynamically constructs a question and conversation history aware context graph at each turn, and then applies a flow mechanism to process a sequence of context graphs.
The {\em Prediction Layer} predicts the answers based on the matching score of the question embedding and the context graph embedding.
The details of these modules are given next.

\subsection{Encoding Layer}

We apply an effective encoding layer to encode the context and the question, which additionally exploits conversation history and interactions between them.

\paragraph{Linguistic features.}
For context word $c_j$, we encode linguistic features into a vector $f_{\text{ling}}(c^{(i)}_{j})$ concatenating POS (part-of-speech), NER (named entity recognition) and exact matching (which indicates whether $c_j$ appears in $Q^{(i)}$) embeddings.

\paragraph{Pretrained word embeddings.}
We use 300-dim GloVe \cite{pennington2014glove} embeddings and 1024-dim BERT \cite{devlin2018bert} embeddings to embed each word in the context and the question.
Compared to GloVe, BERT better utilizes contextual information when embedding words.

\paragraph{Aligned question embeddings.}
Exact matching matches words on the surface form; we further apply an attention mechanism to learn soft alignment between context words and question words.
Since this soft alignment operation is conducted in parallel at each turn,
for the sake of simplicity, we omit the turn index $i$ when formulating the alignment operation.
Following Lee et al.~\shortcite{lee2016learning}, for context word $c_j$ at each turn, we incorporate an aligned question embedding
\begin{equation}
\begin{aligned}
f_{align}(c_{j})=\sum_{k}{\beta_{j,k} \vec{g}_{k}^{Q}}
\end{aligned}
\end{equation}
where $\vec{g}_{k}^{Q}$ is the GloVe embedding of the $k$-th question word $q_k$ and $\beta_{j,k}$ is an attention score between context word $c_j$ and question word $q_k$.
The attention score $\beta_{j,k}$ is computed by
\begin{equation}
\begin{aligned}
\beta_{j,k} \ \propto \ \text{exp}(\text{ReLU}(\vec{W} \vec{g}_{j}^C)^T \text{ReLU}(\vec{W} \vec{g}_{k}^{Q}))
\end{aligned}
\end{equation}
where $\vec{W}$ is a $d \times 300$ trainable weight with
$d$ being the hidden state size, and $\vec{g}_{j}^C$ is the GloVe embedding of context word $c_j$.
To simplify notation, we denote the above attention mechanism as $\text{Align}(\vec{X}, \vec{Y}, \vec{Z})$, meaning that an attention matrix is computed between two sets of vectors $\vec{X}$ and $\vec{Y}$, which is later used to get a linear combination of vector set $\vec{Z}$. Hence we can reformulate the above alignment as
\begin{equation}
\begin{aligned}
f_{\text{align}}(C) = \text{Align}(\vec{g}^C, \vec{g}^{Q}, \vec{g}^{Q})
\end{aligned}
\end{equation}

\paragraph{Conversation history.}
Following Choi et al.~\shortcite{choi2018quac}, we concatenate a feature vector $f_{\text{ans}}(c^{(i)}_{j})$ encoding previous $N$ answer locations to context word embeddings.
Preliminary experiments showed that it is
helpful to also prepend previous $N$ question-answer pairs to a current question.
In addition,
to each word vector in an augmented question,
we concatenate a turn marker embedding $f_{\text{turn}}(q_{k}^{(i)})$ indicating which turn the word belongs to.


In summary, at the $i$-th turn in a conversation, each context word $c_j$ is encoded by a vector $\vec{w}^{(i)}_{c_j}$ which is a concatenation of linguistic vector $f_{\text{ling}}(c^{(i)}_{j})$, word embeddings (i.e., $\vec{g}_j^C$ and $\text{BERT}_j^C$),
aligned vector $f_{\text{align}}(c^{(i)}_{j})$ and answer vector $f_{\text{ans}}(c^{(i)}_{j})$. And each question word $q_{k}^{(i)}$ is encoded by a vector $\vec{w}_{q_k}^{(i)}$ which is a concatenation of
word embeddings (i.e., $\vec{g}_{k}^{Q^{(i)}}$ and $\text{BERT}_{k}^{Q^{(i)}}$)
and turn marker vector $f_{\text{turn}}(q_{k}^{(i)})$.
We denote $\vec{W}^{(i)}_C$ and $\vec{W}^{(i)}_Q$ as a sequence of context word vectors $\vec{w}^{(i)}_{c_j}$ and question word vectors $\vec{w}_{q_k}^{(i)}$, respectively.

\subsection{Reasoning Layer}
When performing reasoning over context, unlike most previous methods that regard context as a word sequence, we opt to treat context as a ``graph" of words that
captures rich semantic relationships among words, and apply a Recurrent Graph Neural Network to process a sequence of context graphs.

\subsubsection{Question Understanding}
For a question $Q^{(i)}$, we apply a bidirectional LSTM \cite{hochreiter1997long} to the question embeddings $\vec{W}^{(i)}_Q$ to obtain contextualized embeddings $\vec{Q}^{(i)} \in \mathbb{R}^{d \times n}$.
\begin{equation}
\begin{aligned}
\vec{Q}^{(i)} = \vec{q}^{(i)}_{1},...,\vec{q}^{(i)}_{n} = \text{BiLSTM}(\vec{W}^{(i)}_Q)
\end{aligned}
\end{equation}

And the question is then represented as a weighted sum of question word vectors via a self attention mechanism,
\begin{equation}
\begin{aligned}
\widetilde{\vec{q}}^{(i)} = \sum_k{a^{(i)}_{k} \vec{q}^{(i)}_{k}}, \ \ \text{where} \ \
a^{(i)}_{k}\ \propto\ \text{exp}(\vec{w}^T \vec{q}^{(i)}_{k})
\end{aligned}
\end{equation}
where $\vec{w}$ is a $d$-dim trainable weight.

Finally, to capture the dependency among question history, we encode the sequence of questions with a LSTM to generate history-aware question vectors.
\begin{equation}\label{eq:question_emb}
\begin{aligned}
\vec{p}^{(1)},...,\vec{p}^{(T)} &= \text{LSTM}(\widetilde{\vec{q}}^{(1)},...,\widetilde{\vec{q}}^{(T)})
\end{aligned}
\end{equation}
The output hidden states of the LSTM network $\vec{p}^{(1)},...,\vec{p}^{(T)}$ will be used for predicting answers.

\subsubsection{Context Graph Learning}
The intrinsic context graph structure is unfortunately unknown.
Moreover, the context graph structure might vary across different turns by considering the changes of questions and conversation history.
Most existing applications of GNNs~\cite{xu2018exploiting,de2018question,song2018exploring} use ground-truth or manually constructed graphs which have some limitations.
First, the ground-truth graphs are not always available.
Second, errors in manual construction process can be propagated to subsequent modules.
Unlike previous methods, we automatically construct graphs from raw context, which are combined with the rest of the system to make the whole learning system end-to-end trainable.
We dynamically build a question and conversation history aware context graph to model semantic relationships among context words at each turn.

Specifically, we first apply an attention mechanism to the context representations $\vec{W}^{(i)}_C$ (which additionally incorporate both question information and conversation history) at the $i$-th turn to compute an attention matrix $\vec{A}^{(i)}$, serving as a weighted adjacency matrix for the context graph, defined as,
\begin{equation}\label{eq:graph_learner}
\begin{aligned}
\vec{A}^{(i)} = (\vec{W}^{(i)}_C \odot \vec{u})^T \vec{W}^{(i)}_C
\end{aligned}
\end{equation}
where $\odot$ denotes element-wise multiplication, and $\vec{u}$ is a non-negative $d_c$-dim trainable weight vector which
learns to highlight different dimensions of $\vec{w}^{(i)}_{c_j}$ whose dimension is $d_c$.

Considering that a fully connected context graph is not only computationally expensive but also might introduce noise (i.e., unimportant edges),
a simple kNN-style graph sparsification operation is applied to select the most important edges from the fully connected graph, resulting in a sparse graph.
To be concrete, given a learned attention matrix $\vec{A}^{(i)}$,
we only keep the $K$ nearest neighbors (including itself) as well as the associated attention scores (i.e., the remaining attentions scores are masked off) for each context node.
We then apply a softmax function to these selected adjacency matrix elements to get a normalized adjacency matrix.
\begin{equation}\label{eq:graph_knn}
\begin{aligned}
\widetilde{\vec{A}}^{(i)} &= \text{softmax}(\text{topk}(\vec{A}^{(i)}))
\end{aligned}
\end{equation}
Note that the supervision signal is still able to back-propagate through the kNN-style graph sparsification module since the $K$ nearest attention scores are kept and used to compute the weights of the final normalized adjacency matrix.


\subsubsection{Context Graph Reasoning}

\begin{figure}[!htb]
  \centering
    \includegraphics[keepaspectratio=true,scale=0.2]{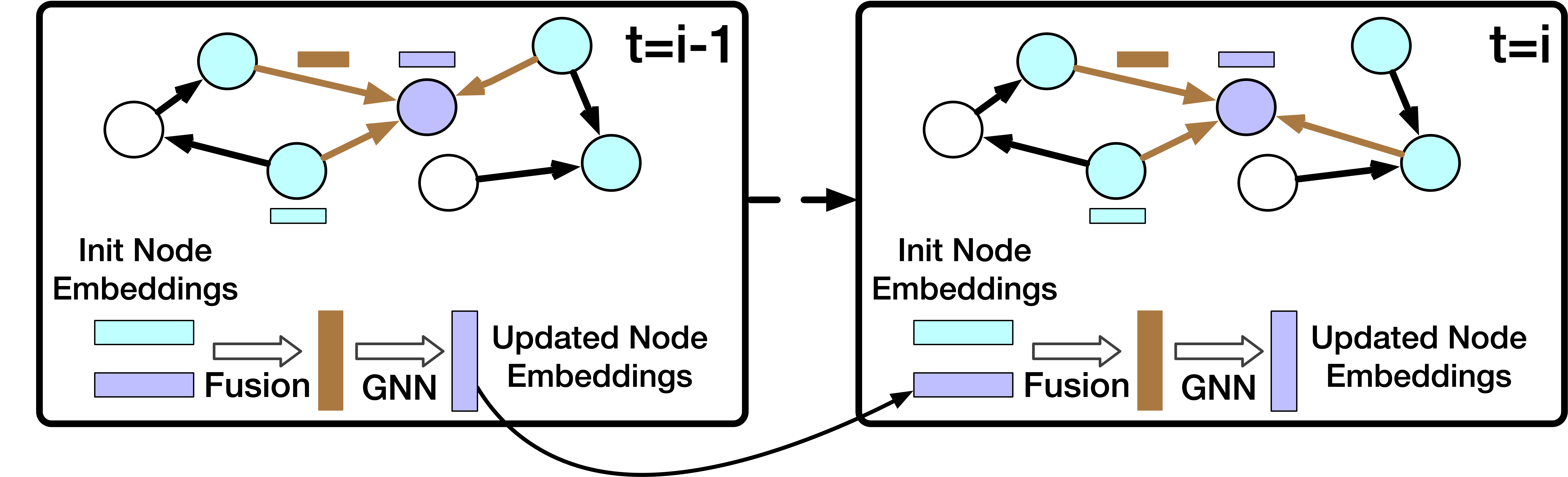}
  \caption{Architecture of the proposed Recurrent Graph Neural Network for processing a sequence of context graphs. }
  \label{fig:rgnn_arch}
\end{figure}

When reasoning over a sequence of context graphs, we want to consider not only the relationships among graph nodes,
but also the sequential dependencies among graphs.
Especially for the conversational MC task, we hope the results of previous reasoning processes can be incorporated into the current reasoning process since they potentially capture important information
for answering the current question.

Therefore,
we propose a novel {\em Recurrent Graph Neural Network} (RGNN) to process a sequence of graphs, as shown in \cref{fig:rgnn_arch}.
As we advance in a sequence of graphs, we process each graph using a shared GNN cell and the GNN output will be used when processing the next graph.
One can think that it is analogous to an RNN-style structure where the main difference is that each element in a sequence is
not a data point, but instead a graph.
Our RGNN module combines the advantages of RNNs which are good at sequential learning (i.e., modeling sequential data), and GNNs which are good at relational reasoning (i.e., modeling graph-structured data).

The computational details of RGNN are as follows.
Let us denote $\vec{C}^{(i)}$ as the initial context node embedding at the $i$-th turn.
Before we apply a GNN to the context graph $\mathcal{G}^{(i)}$, we update its node embeddings by fusing both the original node information $\vec{C}^{(i)}$ and the updated node information $\widebar{\vec{C}}^{(i-1)}$ computed by a parameter-sharing GNN at the $(i-1)$-th turn via a fusion function,
\begin{equation}
\begin{aligned}
\widebar{\vec{C}}^{(i)} = \text{GNN}(\text{Fuse}(\vec{C}^{(i)}, \widebar{\vec{C}}^{(i-1)}), \widetilde{\vec{A}}^{(i)})\\
\end{aligned}
\end{equation}
where we set $\widebar{\vec{C}}^{(0)} = \vec{C}^{0}$ as we do not incorporate any historical information at the first turn.
The fusion function is designed as a gated sum of two information sources,
\begin{equation}\label{eq:fusion}
\begin{aligned}
\text{Fuse}(\vec{a}, \vec{b}) = \vec{z} * \vec{a} + (1-\vec{z}) * \vec{b}\\
\vec{z} = \sigma(\vec{W}_z [\vec{a}; \vec{b}; \vec{a}*\vec{b}; \vec{a}-\vec{b}]+\vec{b}_z)
\end{aligned}
\end{equation}
where $\sigma$ is a sigmoid function and $\vec{z}$ is a gating vector.
As a result, the graph node embedding outputs of the reasoning process at the previous turn are used as a starting state when reasoning at the current turn.

We use Gated Graph Neural Networks (GGNN) \cite{li2015gated} as our GNN cell, but the framework is agnostic to the particular choice of GNN cell.
In GGNN we do multi-hop message passing through a graph to capture long-range dependency where the same set of network parameters are shared at every hop of computation.
At each hop of computation, for every graph node,
we compute an aggregation vector as a weighted average of all its neighboring node embeddings where the weights come from the normalized adjacency matrices $\widetilde{\vec{A}}^{(i)}$.
Then, a Gated Recurrent Unit (GRU)~\cite{cho2014learning} is used to update node embeddings by incorporating the aggregation vectors.
We use the updated node embeddings at the last hop as the final node embeddings.

To simplify notation, we denote the above RGNN module as $\widebar{\vec{C}}^{(i)} = \text{RGNN}(\vec{C}^{(i)}, \widetilde{\vec{A}}^{(i)}), \ i = 1,\ldots, T$ which takes as input a sequence of graph node embeddings $\{\vec{C}^{(i)}\}_{i=1}^{T}$ as well as a sequence of the normalized adjacency matrices $\{\widetilde{\vec{A}}^{(i)}\}_{i=1}^{T}$,
and outputs a sequence of updated graph node embeddings $\{\widebar{\vec{C}}^{(i)}\}_{i=1}^T$.

While a GNN is responsible for modeling global interactions among context words, modeling local interactions between consecutive context words is also important for the task.
Therefore, before feeding the context word representations to a GNN, we first apply a BiLSTM to encode local dependency, that is, $\vec{C}^{(i)} = \text{BiLSTM}(\vec{W}^{(i)}_C)$, and then use the output $\vec{C}^{(i)}$ as the initial context node embedding.

Inspired by recent work \cite{wang2018multi} on modeling the context with different levels of granularity, we choose to apply
stacked RGNN layers where one RGNN layer is applied on low level representations of the context and the second RGNN layer is applied on high level representations of the context.
The output of the second RGNN layer $\{\widetilde{\vec{C}}^{(i)}\}_{i=1}^T$ is the final context representations.
\begin{equation}\label{eq:stacked_rgnn}
\begin{aligned}
\vec{H}^{(i)}_C &= [\widebar{\vec{C}}^{(i)}; \vec{g}^C; \text{BERT}^C]\\
\vec{H}_Q^{(i)} &= [\vec{Q}^{(i)}; \vec{g}^{Q^{(i)}}; \text{BERT}^{Q^{(i)}}]\\
f^2_{\text{align}}(C^{(i)}) &= \text{Align}(\vec{H}^{(i)}_C, \vec{H}_Q^{(i)},  \vec{Q}^{(i)}]\\
\hat{\vec{C}}^{(i)} &= \text{BiLSTM}([\widebar{\vec{C}}^{(i)}; f^2_{\text{align}}(C^{(i)})])\\
\widetilde{\vec{C}}^{(i)} &= \text{RGNN}(\hat{\vec{C}}^{(i)}, \widetilde{\vec{A}}^{(i)}), \ i = 1,\ldots, T\\
\end{aligned}
\end{equation}

\subsection{Prediction Layer}\label{sec:pred_layer}
We predict answer spans by computing the start and end probabilities of the $j$-th context word for the $i$-th question.
For the sake of simplicity, we omit the turn index $i$ when formulating the prediction layer.
The start probability $P^S_j$ is calculated by,
\begin{equation}
\begin{aligned}
P^S_j & \ \propto \ \text{exp}({\widetilde{\vec{c}}_j}^T \vec{W}_S \vec{p})
\end{aligned}
\end{equation}
where $\vec{W}_S$ is a $d \times d$ trainable weight and $\vec{p}$ (turn index omitted) is the question representation obtained in \cref{eq:question_emb}.
Next, $\vec{p}$ is passed to a GRU cell by incorporating context summary and converted to $\widetilde{\vec{p}}$.
\begin{equation}
\begin{aligned}
\widetilde{\vec{p}} &= \text{GRU}(\vec{p}, \sum_{j}{P^S_j \widetilde{\vec{c}}_j})
\end{aligned}
\end{equation}
Then, the end probability $P^E_j$ is calculated by,
\begin{equation}
\begin{aligned}
P^E_j & \ \propto \ \text{exp}({\widetilde{\vec{c}}_j}^T \vec{W}_E \widetilde{\vec{p}})
\end{aligned}
\end{equation}
where $\vec{W}_E$ is a $d \times d$ trainable weight.

\begin{table*}[ht!]
\centering
\scalebox{1}{
\begin{tabular}{lllllllllll}
  \hline
   & \vline & Child. & Liter. & Mid-High. & News & Wiki & Reddit & Science & \vline & Overall\\
  \hline
  \hline
  PGNet \cite{see2017get} &\vline & 49.0 &43.3& 47.5& 47.5& 45.1 &38.6 &38.1& \vline&44.1\\
  DrQA \cite{chen2017reading} &\vline & 46.7& 53.9& 54.1& 57.8 &59.4 &45.0& 51.0&\vline &52.6\\
  DrQA+PGNet \cite{reddy2018coqa} & \vline& 64.2& 63.7& 67.1& 68.3 &71.4 &57.8& 63.1 &\vline&65.1\\
  BiDAF++ \cite{yatskar2018qualitative} & \vline&66.5 & 65.7 & 70.2 & 71.6 & 72.6 & 60.8 & 67.1& \vline& 67.8\\
  \textsc{FlowQA} \cite{huang2018flowqa} & \vline&73.7 & 71.6 & 76.8 & 79.0 & 80.2 & 67.8 & 76.1 &\vline &75.0\\
  Flow [Unpublished] & \vline& \quad\textrm{--} & \quad\textrm{--} & \quad\textrm{--} & \quad\textrm{--} & \quad\textrm{--} & \quad\textrm{--} & \quad\textrm{--} &\vline &75.8\\
  SDNet \cite{zhu2018sdnet} &\vline &75.4 & 73.9 & 77.1 & \textbf{80.3} & \textbf{83.1} & 69.8 & 76.8 & \vline&76.6\\
  \hline
  \textsc{GraphFlow} &\vline& \textbf{77.1}  & \textbf{75.6} &  \textbf{77.5} & 79.1 & 82.5  & \textbf{70.8}  & \textbf{78.4} & \vline& \textbf{77.3}\\ 
  \hline
  Human &\vline& 90.2 &88.4& 89.8& 88.6 &89.9& 86.7& 88.1 & \vline&88.8\\ \hline
\end{tabular}
}
\caption{Model and human performance (\% in F1 score) on CoQA test set.}
\label{table:coqa_results}
\end{table*}

We apply an answer type classifier to handle unanswerable questions and questions whose answers are not text spans in the context.
The probability of the answer type (e.g., ``unknown'', ``yes'' and ``no'') is calculated as follows,
\begin{equation}
\begin{aligned}
P^C = \sigma(f_c(\vec{p}) [f_{\text{mean}}(\widetilde{\vec{C}}); f_{\text{max}}(\widetilde{\vec{C}})]^T)
\end{aligned}
\end{equation}
where $f_c$ is a dense layer which maps a $d$-dim vector to a $(\text{num\_class} \times 2d)$-dim vector.
Further, $\sigma$ is a sigmoid function for binary classification and a softmax function for multi-class classification.
$f_{\text{mean}}(.)$ and $f_{\text{max}}(.)$ denote the average pooling and max pooling operations, respectively.

\subsection{Training and Inference}
The training objective for the $i$-th turn is defined as the cross entropy loss of both text span prediction (if the question requires it) and answer type prediction where the turn index $i$ is omitted for the sake of simplicity,
\begin{equation}
\begin{aligned}
\mathcal{L} = -I^S (log(P^S_{s}) + log(P^E_{e})) + \log P^C_{t}
\end{aligned}
\end{equation}
where $I^S$ indicates whether the question requires answer span prediction, $s$ and $e$ are the ground-truth start and end positions of the span,
and $t$ indicates the ground-truth answer type.

During inference, we first use $P^C$ to predict whether the question requires text span prediction. If yes, we predict the span to be $\hat{s}, \hat{e}$ with maximum $P^S_{\hat{s}}, P^E_{\hat{e}}$ subject to
certain maximum span length threshold.

\section{Experiments}
In this section, we conduct an extensive evaluation of our proposed model against state-of-the-art conversational MC models. We use three popular benchmarks as described below.
The implementation of our model is publicly available at \url{https://github.com/hugochan/GraphFlow}.

\subsection{Datasets, Baselines and Evaluation Metrics}

CoQA~\cite{reddy2018coqa} contains 127k questions with answers, obtained from 8k conversations.
Answers are in free-form and hence are not necessarily text spans from context.
The average length of questions is only 5.5 words.
The average number of turns per dialog is 15.2.
QuAC~\cite{choi2018quac} contains 98k questions with answers, obtained from 13k conversations. All the answers are text spans from context.
The average length of questions is 6.5 and there are on average 7.2 questions per dialog.
DoQA~\cite{camposconversational} contains 7.3k questions with answers, obtained from 1.6k conversations in the cooking domain.
Similar to CoQA, 31.3\% of the answers are not directly extracted from context.

We compare our method with the following baselines:
PGNet \cite{see2017get}, DrQA \cite{chen2017reading}, DrQA+PGNet \cite{reddy2018coqa}, BiDAF++ \cite{yatskar2018qualitative}, \textsc{FlowQA} \cite{huang2018flowqa},
SDNet \cite{zhu2018sdnet},
BERT \cite{devlin2018bert}
and Flow (unpublished).

Following previous works~\cite{huang2018flowqa,zhu2018sdnet},
we use an extractive approach with answer type classifiers on all benchmarks.
The main evaluation metric is F1 score
which is the harmonic mean of precision and recall at word level between the predication and ground truth.
In addition, for QuAC and DoQA, the Human Equivalence Score (HEQ) is used to judge whether a system performs as well as an average human.
HEQ-Q and HEQ-D are model accuracies at question level and dialog level.
Please refer to \cite{reddy2018coqa,choi2018quac} for details of these metrics.

\subsection{Model Settings}\label{sec:model_settings}
The embedding sizes of POS, NER, exact matching and turn marker embeddings are set to 12, 8, 3 and 3, respectively.
Following Zhu et al.~\shortcite{zhu2018sdnet}, we pre-compute BERT embeddings for each word using a weighted sum of BERT layer outputs.
The size of all hidden layers is set to 300.
When constructing context graphs, the neighborhood size is set to 10.
The number of GNN hops is set to 5 for CoQA and DoQA, and 3 for QuAC.
During training, we apply dropout after embedding layers (0.3 for GloVe and 0.4 for BERT) and RNN layers (0.3 for all).
We use Adamax \cite{kingma2014adam} as the optimizer and the learning rate is set to 0.001.
We batch over dialogs and the batch size is set to 1.
When augmenting the current turn with conversation history, we only consider the previous two turns.
All these hyper-parameters are tuned on the development set.

\subsection{Experimental Results}

As shown in \cref{table:coqa_results}, \cref{table:quac_results}, and \cref{table:doqa_results}, our model outperforms or achieves competitive performance compared with various state-of-the-art baselines.
Compared with \textsc{FlowQA} which is also based on the flow idea, our model improves F1 by 2.3\% on CoQA, 0.8\% on QuAC and 2.5\% on DoQA, which demonstrates the superiority of our RGNN based flow mechanism over the IF mechanism.
Compared with SDNet which relies on sophisticated inter-attention and self-attention mechanisms, our model improves F1 by 0.7\% on CoQA.

\begin{table}
\centering
\begin{tabular}{lllll}
\hline
   & \vline & F1 & HEQ-Q & HEQ-D\\
  \hline
  \hline
  BiDAF++ &\vline& 60.1 & 54.8 & 4.0\\
  \textsc{FlowQA} &\vline& 64.1 & 59.6 & \textbf{5.8}\\
  \hline
 \textsc{GraphFlow} &\vline& \textbf{64.9} & \textbf{60.3} & 5.1\\
  \hline
  Human &\vline& 80.8 & 100 & 100\\ \hline
\end{tabular}
\caption{Model and human performance (in \%) on QuAC test set.}
\label{table:quac_results}
\end{table}

\begin{table}
\centering
\begin{tabular}{lllll}
\hline
   & \vline & F1 & HEQ-Q & HEQ-D\\
  \hline
  \hline
 BERT & \vline&41.4 &38.6&4.8\\
 \textsc{FlowQA} &\vline& 42.8 & 35.5 & 5.0\\
  \hline
 \textsc{GraphFlow} &\vline& \textbf{45.3} & \textbf{41.5} & \textbf{5.3} \\
  \hline
  Human &\vline& 86.7 &\quad\textrm{--}  & \quad\textrm{--} \\ \hline
\end{tabular}
\caption{Model and human performance (in \%) on DoQA test set.}
\label{table:doqa_results}
\end{table}

\subsection{Ablation Study and Model Analysis}

We conduct an extensive ablation study to further investigate the performance impact of different components in our model.
Here we briefly describe ablated systems:
\textrm{--} RecurrentConn removes temporal connections between consecutive context graphs,
\textrm{--} RGNN removes the RGNN module,
\textrm{--} kNN removes the kNN-style graph sparsification operation,
\textrm{--} PreQues does not prepend previous questions to the current turn,
\textrm{--} PreAns does not prepend previous answers to the current turn,
\textrm{--} PreAnsLoc does not mark previous answer locations in the context,
and
\textrm{--} BERT removes pretrained BERT embeddings.
We also show the model performance with no conversation history \textsc{GraphFlow} (0-His) or one previous turn of the conversation history \textsc{GraphFlow} (1-His).

\begin{table}
\centering
\begin{tabular}{lll}
\hline
   & \vline & F1 \\
  \hline
  \hline
 \textsc{GraphFlow} (2-His) & \vline & \textbf{78.3} \\
\qquad \textrm{--} PreQues &\vline & 78.2 \\
\qquad \textrm{--} PreAns &\vline& 77.7  \\
\qquad \textrm{--} PreAnsLoc &\vline& 76.6 \\
\qquad \textrm{--} BERT &\vline& 76.0  \\
\qquad \qquad \textrm{--} RecurrentConn &\vline& 69.9 \\
\qquad \qquad \textrm{--} RGNN &\vline& 68.8 \\
\qquad \qquad \textrm{--} kNN &\vline& 69.9 \\
 \textsc{GraphFlow} (1-His) &\vline& 78.2  \\
 \textsc{GraphFlow} (0-His) &\vline& 76.7 \\ \hline
\end{tabular}
\caption{Ablation study (in \%) on CoQA dev.\ set.}
\label{table:ablation_results}
\end{table}

\cref{table:ablation_results} shows the contributions of the above components on the CoQA development set.
Our proposed RGNN module contributes significantly to the model performance (i.e., improves F1 score by 7.2\%).
In addition, within the RGNN module, both the GNN part (i.e., 1.1\% F1) and the temporal connection part (i.e., 6.1\% F1) contribute to the results.
This verifies the effectiveness of representing a passage as a graph and modeling the temporal dependencies in a sequence of context graphs.
The kNN-style graph sparsification operation also contributes significantly to the model performance.
We notice that explicitly adding conversation history to the current turn helps the model performance.
We can see that the previous answer information is more crucial than the previous question information.
And among many ways to use the previous answer information, directly marking previous answer locations seems to be the most effective.
Last but not least, we find that the pretrained BERT embedding has significant impact on the performance, which demonstrates the power of large-scale pretrained language models.

\subsection{Interpretability Analysis}
\begin{figure}
    \includegraphics[keepaspectratio=true,scale=0.28]{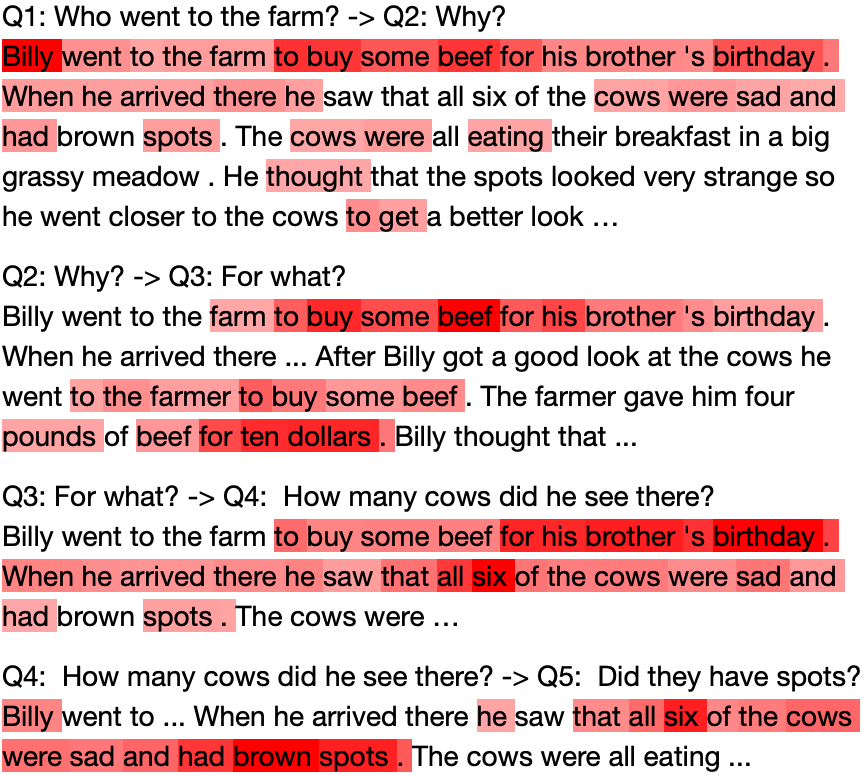}
    \caption{The highlighted part of the context indicates GraphFlow's focus shifts between consecutive question turns.}
    \label{fig:vis_qa}
\end{figure}

Following Huang et al.~\shortcite{huang2018flowqa},
we visualize the changes of hidden representations of context words between consecutive turns.
Specifically, we compute cosine similarity of hidden representations of the same context words at consecutive turns, and then highlight the words that have small cosine similarity scores (i.e., change more significantly).
\cref{fig:vis_qa} highlights the most changing context words (due to the page limit, we do not show full context) between consecutive turns in a conversation from the CoQA dev.\ set.
As we can see, the hidden representations of context words which are relevant to the consecutive questions are changing most and thus highlighted most. We suspect this is in part because when the focus shifts, the model finds out that the context chunks relevant to the previous turn become less important but those relevant to the current turn become more important. Therefore, the memory updates in these regions are the most active.

\section{Conclusion}
We proposed a novel Graph Neural Network (GNN) based model, namely \textsc{GraphFlow}, for conversational machine comprehension (MC) which carries over the reasoning output throughout a conversation.
Besides, we proposed a simple yet effective graph structure learning technique to dynamically construct a question and conversation history aware context graph at each conversation turn.
On three recently released conversational MC benchmarks, our proposed model achieves competitive results compared with previous approaches.
Interpretability analysis shows that our model can offer good interpretability for the reasoning process.
In the future, we would like to investigate more effective ways of automatically learning graph structures from free text and modeling temporal connections between sequential graphs.

\clearpage
\small
\bibliographystyle{named}
\bibliography{ijcai20}

\end{document}